\documentclass[letterpaper, 10 pt, conference]{ieeeconf}  

\IEEEoverridecommandlockouts                              

\usepackage{cite}
\usepackage{graphicx}
\usepackage{epsfig}   
\usepackage{amsmath}
\usepackage{amssymb}
\usepackage{booktabs}
\usepackage{capt-of} 
\usepackage[hidelinks]{hyperref}

\newcommand{\method}{ActMem-VLA}

\usepackage{xcolor}

\usepackage{booktabs}

\title{\LARGE \bf
ActMem-VLA: Action-History Memory with Dual-Expert Denoising for Long-Horizon Vision-Language-Action Policies
}

\author{
Yaxin Zhao$^{1,2}$, Dianye Huang$^{2,\dagger}$, Chenwei Wang$^{2}$, Chenguang Yang$^{3}$, and Zhongliang Jiang$^{2}$
\thanks{
$^\dagger$ Corresponding author (email: dianye.huang@hku.hk)
}
\thanks{$^1$ Harbin Institute of Technology, Harbin, China.}
\thanks{$^2$
Medical Intelligence and Robotic Cognition (MIRoC) Lab, Department of Mechanical Engineering, The University of Hong Kong (HKU), Hong Kong SAR, China.
}
\thanks{$^3$ Department of Computing, The Hong Kong Polytechnic University, Hong Kong SAR, China}
}

\makeatletter
\let\phase@originalmaketitle\@maketitle

\def\@maketitle{%
\phase@originalmaketitle

\noindent
\begin{minipage}{\textwidth}
    \vspace{15pt}
    \centering
    \includegraphics[width=0.98\linewidth]{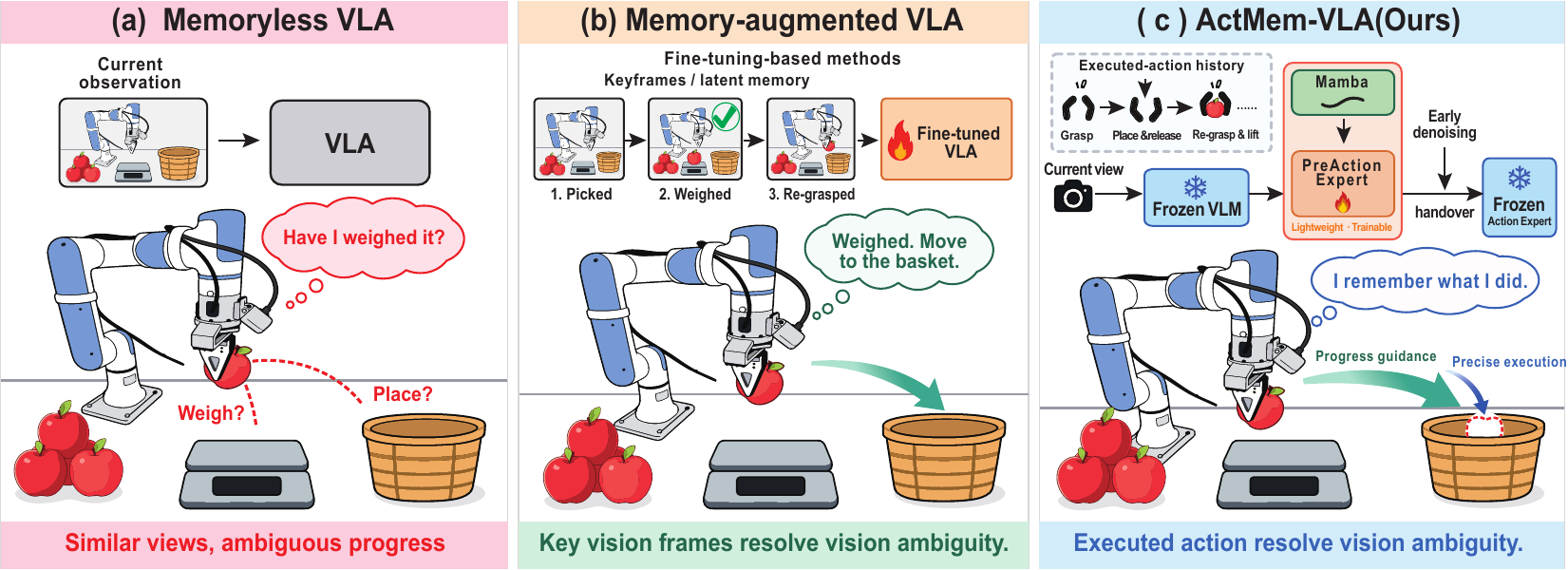}
    \captionof{figure}{%
        \textbf{Motivation: similar observations can require different actions depending on task progress.} In the apple-weighing task, holding the apple above the scale occurs both before weighing and after re-grasping, despite requiring different subsequent actions. (a) A memoryless VLA lacks the history needed to distinguish these stages. (b) Representative memory-augmented approaches that fine-tune the base VLA incorporate historical context to resolve the ambiguity. (c) \method{} instead uses executed-action history as compact progress-aware memory to steer early denoising through a lightweight PreAction Expert, before handing the intermediate action state to the frozen Action Expert for refinement.       
        }
    \label{fig:teaser}
\end{minipage}
}
\makeatother

\begin{document}

\maketitle
\setcounter{figure}{1}

\begin{abstract}
Vision-language-action (VLA) models have driven rapid progress in robotic manipulation, demonstrating strong fine-grained control and promising performance on long-horizon tasks. However, many existing VLAs lack explicit access to interaction history, making them vulnerable to perceptual aliasing: similar current observations and robot states at different task stages may induce action ambiguity and lower success rate. Existing methods incorporate temporal or progress cues through feature conditioning, action-prior modification, or sampling guidance. However, methods that jointly fine-tune memory modules and the base VLA incur additional policy-training costs, motivating the separation of trainable history-conditioned steering from frozen base-policy refinement. We propose ActMem-VLA, a dual-expert handover architecture that augments a frozen, fine-tuned VLA with a memory plugin comprising a Mamba-based memory module and a lightweight PreAction Expert (PAE). Specifically, Mamba encodes executed-action history into memory that conditions PAE alongside current context. With these inputs, PAE steers task progression during early, high-noise denoising, then passes the partially denoised action to the frozen Action Expert (AE) to refine action details during the remaining low-noise steps. The fine-tuned base VLA remains frozen throughout training, while only the Mamba module and PAE are jointly optimized. On LIBERO-Mem, ActMem-VLA achieves 80.8\% average success across all ten tasks, compared with 65.2\% for $\pi_{0.5}$ and 49.5\% for MemoryVLA, while introducing only 3.45\% additional parameters. Across four real-world tasks, it improves the average success rate over $\pi_{0.5}$ by 28.8\%.

\end{abstract}

\begin{figure*}[t]
    \centering
    \includegraphics[width=0.75\textwidth]{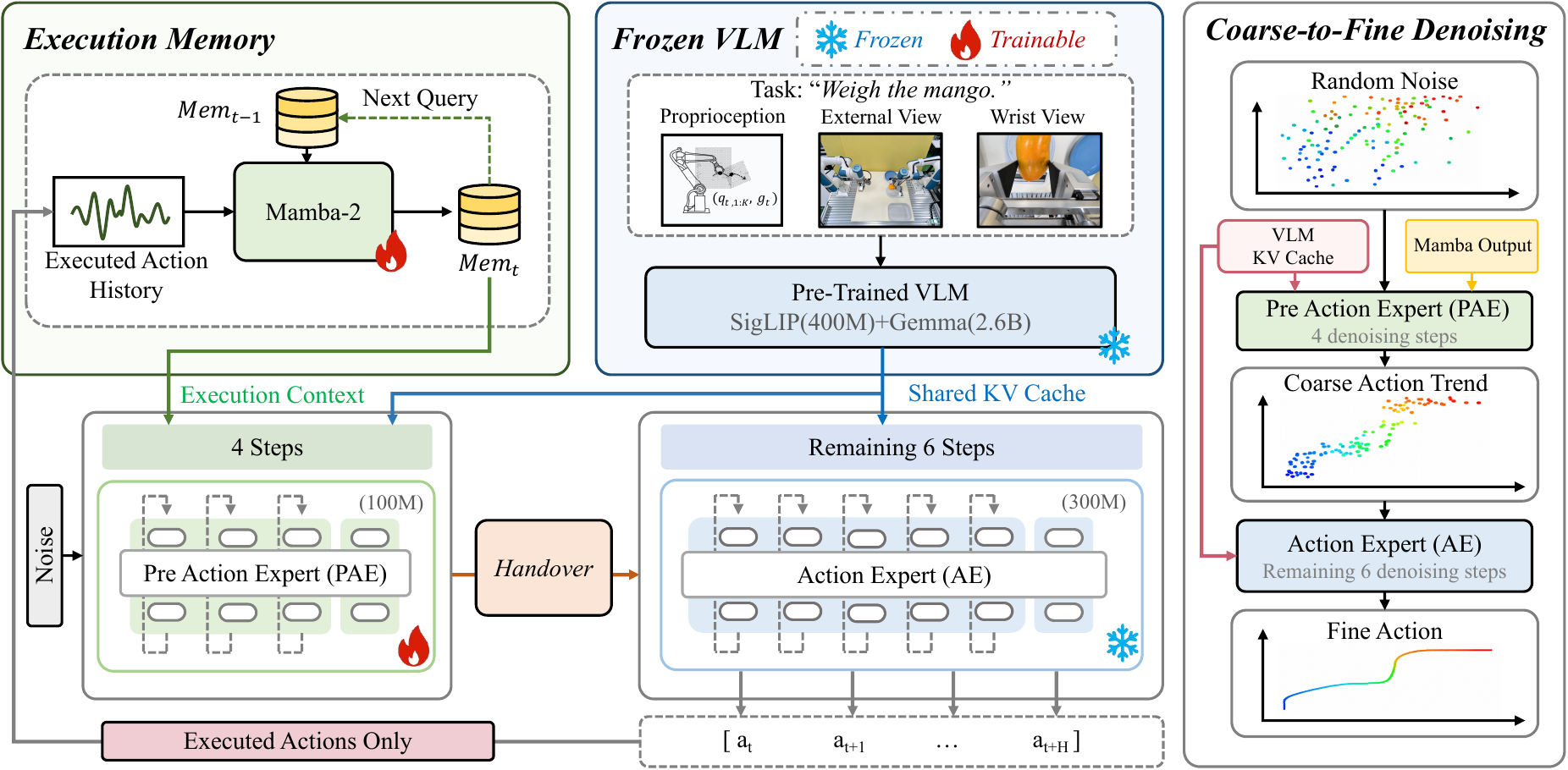}
    \caption{
    \textbf{Overview of \method{}.}
        The Execution Memory module encodes previously executed actions into a compact execution context. Conditioned on this context and the current observation, the PreAction Expert performs early, high-noise denoising. The partially denoised action state is then handed to the frozen Action Expert, which performs the remaining low-noise steps to produce the action chunk. It is noted that only Mamba and the lightweight PreAction Expert are optimized; the base VLM and original Action Expert remain frozen. 
    }
    \label{fig:framework}
\end{figure*}

\section{Introduction}
\label{sec:introduction}
\par
Vision-language-action (VLA) models~\cite{openvla,openvlaoft} combine vision-language representations with robot action generation, enabling language-conditioned manipulation. Recent flow-matching VLAs~\cite{pi0,pi05,pi07,gr00t} further demonstrate strong fine-grained control and promising performance on long-horizon tasks. However, reliable execution of such tasks requires not only accurate control but also awareness of what has already been done~\cite{contextvla,global,liberomem}. Many existing VLAs~\cite{xiaomi,holobrain,pi05} generate actions based solely on the current observation and robot state, without explicitly incorporating interaction history. Consequently, when visually similar observations recur at different stages of a task~\cite{ambiguity,rmbench}, the model may be unable to distinguish the current task progress, creating ambiguity in determining the appropriate next action for continued execution.

\par
Consider a robot tasked with picking an apple from a pile, weighing it on a scale, and then placing it in a basket (Fig.~\ref{fig:teaser}). At two different stages, the robot may find itself holding the apple just above the scale: once before weighing it and again after weighing is complete. Although these two situations can appear nearly identical in the current observation and robot state, they require opposite next actions—the robot should first lower the apple onto the scale, but later move it away toward the basket. Without knowing what has already happened, the robot cannot reliably distinguish between these two stages. This \emph{perceptual aliasing} can result in repeated actions, skipped steps, or hesitation during long-horizon execution. 


\par
To address this challenge, recent studies have explored incorporating historical context into action generation, as insufficient temporal awareness can weaken dependencies between consecutive actions, resulting in fragmented trajectories and reduced task-level coherence. Recent efforts mainly incorporate interaction history in three ways: directly conditioning action generation on historical observations~\cite{contextvla,eventvla,kemo,mem,retrieval}, storing and retrieving task-relevant information through memory-conditioned features~\cite{pointtrack,duallatent,VQ,cyclemanip}, or recursively compressing past interactions into recurrent latent representations~\cite{memoryvla,remem,chainvla,single_mamba,tacmamba}. These strategies respectively preserve history explicitly, selectively retain it through external memory, or implicitly summarize it into an evolving internal state.

\par
Directly incorporating historical observations suffers from two major limitations. First, processing multiple past frames substantially increases computational cost and inference latency, hindering real-time control. Second, historical frames contain considerable redundancy, both across temporally adjacent frames and among static pixels within each frame, which can obscure the task-relevant changes that actually characterize interaction progress. To represent history more precisely and efficiently, recent approaches therefore focus on compact dynamic information rather than raw visual observations. HiF-VLA~\cite{hif} uses inter-frame motion as a compact proxy for interaction dynamics, capturing state changes while discarding redundant static content, whereas Chen~\emph{et al.}~\cite{pointtrack} adopts an object-centric representation that tracks task-relevant points over time to summarize historical interactions. Wang~\emph{et al.}~\cite{wang2026remember}, further demonstrates the value of action history for resolving partial observability by learning compact behavioral memory through past-action reconstruction to condition a diffusion policy.

\par
Despite these advances in efficient history representation, existing methods typically inject historical information into the policy input or internal features, tightly coupling memory modeling with action generation. As a result, equipping an already fine-tuned VLA with memory generally requires additional policy training to jointly adapt the memory and action-generation components~\cite{eventvla,wang2026nativemem,memoryvla,streamvla}. This coupling limits the flexibility of extending existing VLAs with memory, particularly when retraining the action expert is costly or undesirable. In contrast, recent approaches show that a frozen policy can be steered through separately learned guidance~\cite{global,dynaguide}, opening an alternative pathway for incorporating historical context without modifying the underlying action expert. Motivated by this paradigm, we introduce a lightweight memory plugin that learns to extract guidance from interaction history and steer action generation, while keeping the pretrained action expert entirely frozen.


\par
In this study, we propose \method{}, a dual-expert handover architecture that augments a frozen, fine-tuned VLA with a Mamba-based memory module and a lightweight PreAction Expert (PAE), as illustrated in Fig.~\ref{fig:framework}. 
We construct memory solely from executed actions, as mixing high-dimensional visual observations with low-dimensional actions can lead to modality imbalance, allowing redundant visual content to dominate the memory representation and obscure the action dynamics that directly reflect task progress. The Mamba module encodes the history of executed actions into a compact representation of task progress, which, together with the current context, conditions the PAE. During action generation, the PAE handles the early, high-noise denoising steps, steering the generative trajectory toward actions consistent with the interaction history. The partially denoised action state is then handed over to the original, frozen Action Expert (AE), which completes the remaining low-noise steps for fine-grained action refinement. This dual-expert handover decouples trainable, history-aware steering from frozen policy refinement within a single denoising trajectory, enabling temporal adaptation without modifying the base action expert. Practically, \method{} provides a \textbf{plug-and-play mechanism} for equipping existing VLAs with temporal memory, enabling parameter-efficient adaptation to new tasks while retaining their previously learned action-generation capabilities.

\par
This design leverages the executed-action history as a compact source of task-progress information while confining all adaptation to the memory plugin. The fine-tuned base VLA remains entirely frozen during training, with only the Mamba memory module and PAE jointly optimized. In our implementation, the plugin introduces only $3.45\%$ additional parameters relative to the base VLA, providing history-aware action generation with minimal trainable overhead. The main contributions of this work are summarized as follows:
\begin{enumerate}
    \item We introduce an \textbf{executed-action memory} mechanism that encodes interaction history into compact, progress-relevant context, enabling a frozen, fine-tuned VLA to resolve perceptual aliasing without relying on redundant historical visual observations.
    \item We develop a \textbf{dual-expert handover} mechanism that uses a lightweight, memory-conditioned PAE for early history-aware denoising and the frozen AE for subsequent fine-grained refinement, enabling plug-and-play temporal adaptation with only $3.45\%$ additional parameters relative to the base VLA.
\end{enumerate}
To evaluate the effectiveness of the proposed \method{}, we conduct experiments in both simulated and real-world settings. On LIBERO-Mem, \method{} achieves an average success rate of 80.8\% across all ten tasks, substantially outperforming $\pi_{0.5}$ (65.2\%) and MemoryVLA (49.5\%). Across four real-world tasks, \method{} further improves the average success rate over $\pi_{0.5}$ by 28.8\%, demonstrating its effectiveness in solving perceptual aliasing in robotic manipulation.

\section{Related Work}
\label{sec:related_work}

\subsection{Vision-Language-Action Policies}
\par
Vision-language-action (VLA) models combine visual and language representations with robot action generation~\cite{openvla,openvlaoft}. Flow-matching VLAs~\cite{pi0,pi05,pi07,gr00t} generate continuous action chunks through iterative denoising. Policies without explicit historical context can encounter ambiguity when similar current inputs recur at different task stages~\cite{liberomem,robomme,rmbench,robomemarena}.
Our work builds on an already fine-tuned flow-matching
VLA~\cite{pi05} and introduces history-conditioned early denoising
followed by a handover to the original action expert (AE).
Both the vision-language model (VLM) and original AE
remain frozen during this adaptation.

\subsection{Memory Mechanisms in Robotic Policies}
\par
Temporal memory helps robotic policies resolve ambiguities that cannot be determined from the current observation alone. Existing methods preserve historical information through multi-frame observations, salient events, retrieval, or compact latent memories~\cite{contextvla,eventvla,kemo,retrieval,remem,mem,VQ,duallatent}. Recurrent mechanisms further compress interaction history across successive decisions~\cite{remem,single_mamba,tacmamba,chainvla}. CAMP~\cite{wang2026remember} learns compact behavioral memory from historical observation-action pairs through past-action reconstruction pretraining and uses this memory to condition a diffusion policy for partially observable manipulation.
Since the visual content of past observations is largely redundant, costly to process, and can itself introduce ambiguity, \method{} instead uses the robot's \emph{executed-action history} as a progress-relevant signal. A lightweight Mamba~\cite{mamba} module compresses previous actions into memory that conditions the PreAction Expert during early denoising, while the original VLM and AE remain frozen. 

\subsection{Structured Denoising for Action Generation}
\par
Recent studies increasingly exploit the internal structure of iterative action generation~\cite{ambient,RTC}. Different denoising stages can capture different levels of action information, motivating noise-dependent training or guidance~\cite{ambient,dynaguide,global}. In contrast, \method{} partitions a single denoising trajectory across two experts: a memory-conditioned PAE handles early high-noise steps before handing the intermediate action state to the frozen AE for low-noise refinement. This design enables history-aware steering at the stage where the overall action trajectory is formed, while preserving the fine-grained control capability of the pretrained action expert.


\begin{figure*}[t]
\centering
\begin{minipage}[c]{0.71\textwidth}
  \centering
  \includegraphics[width=\linewidth]{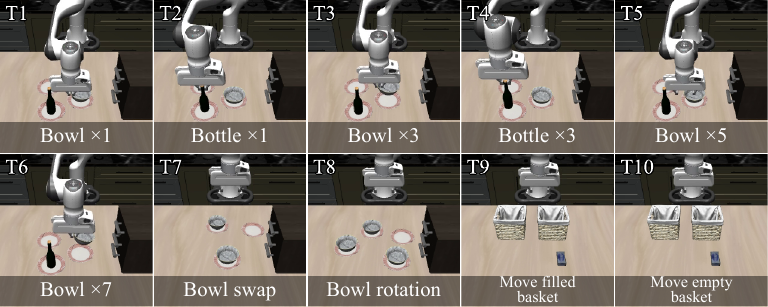}
\end{minipage}%
\hfill
\begin{minipage}[c]{0.26\textwidth}
  \centering
  \includegraphics[width=\linewidth]{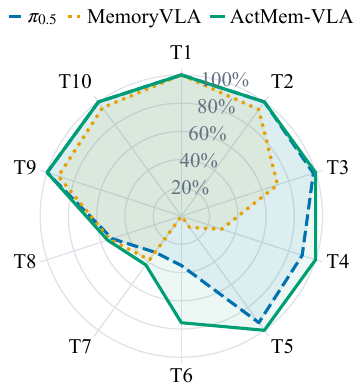}
\end{minipage}

\caption{\textit{Left:} Ten simulation tasks in LIBERO-Mem.
\textit{Right:} Success rates of different methods across these tasks.}
\label{fig:simulation_overview}
\end{figure*}

\begin{figure*}[t!]
\centering
\includegraphics[width=\textwidth]{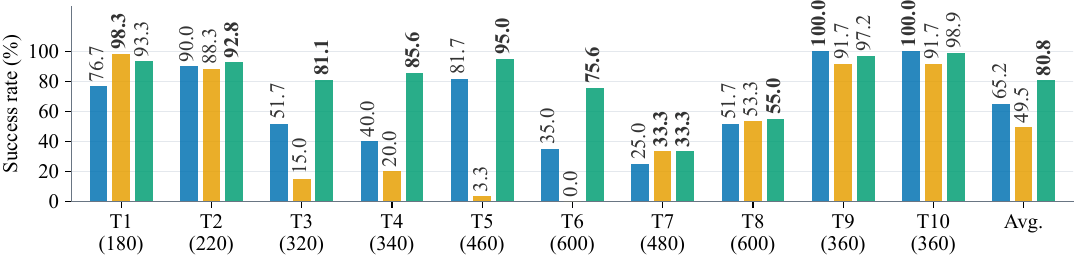}
\caption{Simulation comparison results on LIBERO-Mem. Success rates under task-specific step limits on \texttt{T1}-\texttt{T10}. The number in parentheses below each task indicates the step limit. Episodes that do not complete the task within this limit are counted as failures.}
\label{fig:execution_efficiency}
\end{figure*}

\section{Method}
\label{sec:method}
\subsection{Problem Description}
\label{sec:method_overview}

\par
For a given language instruction, let $\pi(\mathbf{A}_t \mid \mathbf{o}_t)$ denote an action-chunking policy conditioned on the current observation $\mathbf{o}_t$. At each policy query, it predicts an action chunk $\mathbf{A}_t = [\mathbf{a}_t,\mathbf{a}_{t+1},\ldots,\mathbf{a}_{t+H-1}]$ with prediction horizon $H$. The first $s \leq H$ actions are executed before the next query, where $s$ is the execution horizon. We consider flow-matching VLAs, such as $\pi_{0.5}$~\cite{pi05}, whose action expert (AE) transforms Gaussian noise $\mathbf{A}_t^0 \sim \mathcal{N}(0,I)$ into an action chunk using a learned velocity field $v_\pi$. With flow time increasing from $\tau=0$ (noise) to $\tau=1$ (actions), generation proceeds through $n$ Euler updates:
\begin{equation}
    \mathbf{A}_t^{\tau+\frac{1}{n}} = \mathbf{A}_t^\tau + 
    v_\pi(\mathbf{A}_t^\tau,\mathbf{o}_t,\tau)/n,
    \label{eq:base_flow}
\end{equation}
where $\tau \in \{0,1/n,\ldots,(n-1)/n\}$ and $\mathbf{A}_t^1 = \mathbf{A}_t$.

\par
In manipulation tasks that are non-Markovian with respect to the current observation, similar observations $\mathbf{o}_t \approx \mathbf{o}_{t'}$ may occur at different task stages yet require different subsequent actions. Let $\mathcal{H}_t$ denote the action history executed before time $t$. Although $\mathcal{H}_t$ and $\mathcal{H}_{t'}$ can provide different evidence of task progress, the memoryless policy, such as $\pi_{0.5}$, does not condition on these histories. Consequently, it may select stage-inconsistent actions or alternate between competing behaviors across successive queries.

\par
As illustrated in Fig.~\ref{fig:framework}, ActMem-VLA addresses this ambiguity by augmenting a frozen flow-matching VLA with an Execution Memory module and a lightweight PreAction Expert (PAE). The memory module encodes $\mathcal{H}_t$ into a memory representation $\mathbf{m}_t$, which conditions the PAE together with the current observation context. For a \textit{handover point $r \in (0,1)$}, the PAE performs early denoising over $\tau \in [0,r)$. The frozen AE then continues from the resulting intermediate action state to $\tau=1$, without reinitializing noise. This introduces execution-history conditioning into the same $n$-step denoising process while retaining the original AE for subsequent refinement. Only the execution memory module and PAE are trained; the base vision-language model (VLM) and AE remain frozen.

\subsection{Mamba-Based Execution Memory}
\label{sec:method_memory}
\par
We use an $N$-layer Mamba-2 module~\cite{mamba} with hidden width $d_m$ to encode executed-action history. At time $t$, the most recently executed $s$ actions form the block $\mathbf{A}_{t-s}^{\mathrm{exec}} =[\mathbf{a}_{t-s},\ldots,\mathbf{a}_{t-1}]$.
Given the recurrent state $\mathbf{S}_{t-s}$, the memory is updated as
\begin{equation}
    (\mathbf{m}_t,\mathbf{S}_t)
    =
    \operatorname{Mamba2}\!\left(
        \mathbf{A}_{t-s}^{\mathrm{exec}},
        \mathbf{S}_{t-s}
    \right),
    \label{eq:history_memory}
\end{equation}
where $\mathbf{S}_t$ is the updated recurrent state and $\mathbf{m}_t\in\mathbb{R}^{d_m}$ is the memory representation at time $t$. We project $\mathbf{m}_t$ into history keys and values that condition the PAE during early denoising.

\subsection{PreAction Expert and Denoising Handover}
\label{sec:method_pe}
\par
We initialize the lightweight PAE with $P$ Transformer blocks copied from selected layers of the fine-tuned AE. The original AE remains frozen. Each PAE block receives the history KV derived from $\mathbf{m}_t$ and the prefix KV produced by the corresponding frozen VLM layer when processing the current observation and instruction\cite{shallowpi}.

\par
During inference, the PAE predicts the velocity field $v_\pi$ and performs denoising updates using Eq.~\eqref{eq:base_flow} over $0 \leq \tau < r$. At $\tau=r$, the intermediate action state $\mathbf{A}_t^r$ is passed directly to the frozen AE, which continues denoising to $\tau=1$.

\subsection{Joint Training and Efficient Implementation}
\label{sec:method_training}
\par
We jointly train the memory module and PAE on demonstration episodes while keeping the VLM and AE frozen. Within each episode, $\mathbf{m}_t$ is computed from actions preceding time $t$. Memory updates follow the same $s$-action blocks as inference, with padding for incomplete blocks.

\par
For each training query, we sample a flow time $\tau\in[0,r]$ and Gaussian noise
$\boldsymbol{\epsilon}\sim\mathcal{N}(0,I)$, and construct noisy actions $\mathbf{A}_t^\tau =(1-\tau)\boldsymbol{\epsilon}+\tau\mathbf{A}_t$. Conditioned on the current observation context and $\mathbf{m}_t$, the PAE predicts the velocity field $v_\pi$. The memory module and PAE are optimized using the flow-matching loss:
\begin{equation}
    \mathcal{L}_{\mathrm{flow}}
    =
    \mathbb{E}_{t,\tau,\boldsymbol{\epsilon}}
    \left[
        \left\|
        v_\pi(\mathbf{A}_t^\tau,\mathbf{o}_t,
              \mathbf{m}_t,\tau)
        -(\mathbf{A}_t-\boldsymbol{\epsilon})
        \right\|_2^2
    \right],
    \label{eq:flow_objective}
\end{equation}
averaged over valid action positions. To accelerate training, the selected prefix KV from the frozen VLM can be precomputed and reused.




\begin{figure*}[t!]
\centering
\includegraphics[width=\textwidth]{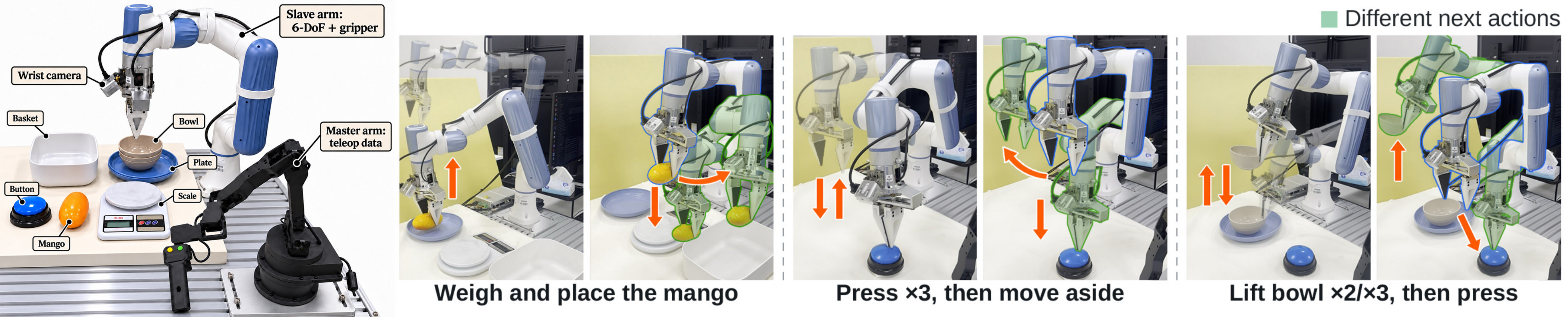}
\caption{Overview of the hardware setup and representative execution sequences for the four real-world evaluation
tasks.}
\label{fig:real_world_setup}
\end{figure*}

\section{EXPERIMENTS}
\label{sec:experiments}
\par
In this section, we evaluated ActMem-VLA across ten simulation and four real-world tasks. Ablation studies were also conducted to assess the contributions of the proposed components and investigate the effects of key hyperparameter choices.     

\subsection{Experimental Settings}
\label{sec:exp}
\par
\noindent\textbf{Simulation setup.}
As illustrated in Fig.~\ref{fig:simulation_overview}, we evaluate ActMem-VLA on LIBERO-Mem~\cite{liberomem}, a simulation benchmark comprising ten memory-dependent manipulation tasks denoted as \texttt{T1}-\texttt{T10}. All simulation experiments and model training were conducted on a computing cluster equipped with eight NVIDIA H100 GPUs.

\par
\noindent\textbf{Real-world setup.}
For the real-world experiments, we employed a DOBOT Nova robotic arm equipped with wrist- and head-mounted cameras. The robot was controlled through the ROS framework running on a workstation equipped with an NVIDIA RTX5090 GPU. For each task, we collected 50 training demonstrations via the GELLO~\cite{gello} leader-follower teleoperation device. As illustrated in Fig.~\ref{fig:real_world_setup}, the robotic arm was tasked with four real-world tasks, denoted as \texttt{R1}-\texttt{R4}: 
\begin{enumerate}
    \item \texttt{R1: Two-Cycle Bowl Return.} Lifting the bowl and returning it to the plate twice, then pressing the button once;
    \item \texttt{R2: Mango Weighing.} Picking up the mango, weighing it on the scale, and placing it in the basket;
    \item \texttt{R3: Triple Button Press.} Pressing the button three times, then moving the arm aside;
    \item \texttt{R4: Three-Cycle Bowl Return.} Performing three bowl lift-and-return cycles, then pressing the button once;
\end{enumerate}
where \texttt{R2} evaluates whether the policy can distinguish different task stages when similar robot-object configurations recur before and after weighing. In contrast, \texttt{R1}, \texttt{R3}, and \texttt{R4} focus on repetition-dependent task progress with a distinct terminal action, requiring the policy to both complete the prescribed number of repetitions and transition to the subsequent action rather than continuously repeating the same behavior. 

\par
\noindent\textbf{Baselines.}
To validate the effectiveness of ActMem-VLA in leveraging historical action context, we compared it against two baselines: \textit{(i) Memory-free $\pi_{0.5}$}~\cite{pi05}, which conditions action prediction solely on the current observation, and \textit{(ii) Memory-augmented MemoryVLA}~\cite{memoryvla}, which maintains a memory bank of historical observations and retrieves task-relevant information to condition action generation. In simulation, both baselines were fine-tuned on the full LIBERO-Mem dataset. For the real-world experiments, each baseline was fine-tuned separately using the training demonstrations collected for the corresponding task.

\par
\noindent\textbf{Evaluation Protocol.}
For the simulation comparison, we evaluate all three methods. Each method is trained with three random seeds, and each resulting checkpoint is evaluated using three evaluation seeds. For each evaluation seed, we conduct 20 trials per task, resulting in 180 trials per task for each method. For the ablation study, we use a single training seed and three evaluation seeds, with 20 trials per task for each seed. In the real-world experiments, we compare $\pi_{0.5}$ with \method{}, with each method evaluated over 20 trials per task.

\subsection{Simulation Results}
\label{sec:exp-sim}
\subsubsection{Comparison under the LIBERO-Mem 600-Step Episode Limit}
\par
Following the standard LIBERO-Mem evaluation protocol, each episode is allowed up to 600 steps for task completion. As shown in the radar plot of Fig~\ref{fig:simulation_overview}, \method{} achieves comparable or superior performance in success rates against the two baselines, with substantial gains on \texttt{T6}. However, even the memoryless $\pi_{0.5}$ baseline achieves nearly 100\% success on several tasks, for instance, \texttt{T1}-\texttt{T3}, which highlights a limitation of the current LIBERO-Mem task completion criterion: a policy may satisfy the success condition through executing cyclic action sequences. Consequently, success rates under the standard protocol do not reliably reveal whether a policy uses memory to track task progress.

\subsubsection{Comparison under Task-Specific Step Limits}
To account for the substantial variation in task execution lengths, we derive a task-specific step limit from the training-demonstration length distribution of each task.

\par
Let $\mathcal{D}_i$ denote the collection of full training-demonstration lengths for task $i$. We define its step limit as
\begin{equation}
    B_i = \min\!\left(
        600,\;
        20\left\lceil
        \frac{\mathcal{Q}_{0.95}(\mathcal{D}_i)}{20}
        \right\rceil
    \right),
    \label{eq:expert_step_budget}
\end{equation}
where $\mathcal{Q}_{0.95}(\cdot)$ denotes taking the 95th percentile element of the collection. This percentile is rounded upward to the nearest multiple of 20 and capped at 600 steps. All methods use the same limit for a given task. An episode is counted as successful only if it satisfies the original task-success condition within the assigned limit. 

\par
Under task-specific step limits, \method{} achieves an average success rate of 80.8\% across the ten tasks, outperforming $\pi_{0.5}$ and MemoryVLA by 15.6 and 31.3 percentage points, respectively (Fig.~\ref{fig:execution_efficiency}). It achieves the highest success rate on six tasks and ties with MemoryVLA on \texttt{T7}. The clearest gains occur on \texttt{T3}-\texttt{T6}. For example, on \texttt{T4}, \method{} achieves 85.6\% success, compared with 40.0\% for $\pi_{0.5}$ and 20.0\% for MemoryVLA. On \texttt{T5}, it reaches 95.0\%, while MemoryVLA achieves only 3.3\%. The advantage is task-dependent: MemoryVLA performs best on \texttt{T1}, while $\pi_{0.5}$ leads on \texttt{T9} and \texttt{T10}. Overall, these results demonstrate stronger task-completion performance under demonstration-derived task-specific step limits.

\subsection{Ablation Studies} 
\par
We evaluate component contributions and hyperparameter sensitivity on three challenging tasks: \texttt{T6}, performing seven repetitions of picking up a bowl and placing it on a plate; \texttt{T7}, swapping two bowls between their plates using an empty plate; and \texttt{T8}, cyclically shifting three bowls from left to right using an empty plate. These tasks cover repetition tracking and multi-stage object rearrangement.

\subsubsection{Component Ablations} 
\par
We evaluate two variants to examine the contributions of execution memory and the trainable PAE. \textit{(i). ``w/o Execution Memory"} removes Mamba and trains the PAE using only the current context, without executed-action history. The base VLA remains frozen. \textit{(ii). ``w/o PreAction Expert"} removes the PAE and directly conditions the original frozen AE on Mamba memory during the first four denoising steps. The same AE completes the remaining six steps without direct memory conditioning. Only the memory branch is trained. This variant tests whether direct memory conditioning of the frozen AE can replace the trainable early-denoising expert. 

\par
As shown in Table~\ref{tab:component_ablation}, removing execution memory reduces average success from 55.0\% to 40.6\%, whereas removing the PAE reduces it to 53.3\%. Both ablations have their largest effects on the repetition-intensive \texttt{T6} task: success falls from 80.0\% to 38.3\% without memory and to 58.3\% without the PAE. These drops of 41.7 and 21.7 percentage points support the contributions of both executed-action history and a trainable early-denoising expert to this task. The gains on \texttt{T7} are smaller, with the full model achieving 40.0\%, compared with 33.3\% without memory and 38.3\% without the PAE. On \texttt{T8}, however, both ablations outperform the full model: success increases from 45.0\% to 50.0\% without memory and to 63.3\% without the PAE. The component benefits are therefore task-dependent, with particularly strong gains on repetition tracking and mixed results on object rearrangement. Across the three tasks, the full model achieves the highest average success rate among the evaluated component variants, supporting the combined use of execution memory and a trainable early-denoising expert.

\begin{table}[h]
    \centering
    \caption{Component ablations on \texttt{T6}--\texttt{T8}
    (success rate, \%).}
    \label{tab:component_ablation}
    \begin{tabular*}{\columnwidth}{
        @{\extracolsep{\fill}}lcccc@{}
    }
        \toprule
        \textbf{Variant} & \texttt{T6} & \texttt{T7} & \texttt{T8} & \textbf{Avg.} \\
        \midrule
        w/o Execution Memory & 38.3 & 33.3 & \underline{50.0} & 40.6 \\
        w/o PreAction Expert & \underline{58.3} & \underline{38.3} & \textbf{63.3} & \underline{53.3} \\
        \textbf{\method{}} & \textbf{80.0} & \textbf{40.0} & 45.0 & \textbf{55.0} \\
        \bottomrule
    \end{tabular*}

    \vspace{3pt}
    \parbox{\columnwidth}{
        \scriptsize
        Each ablation removes the indicated component from the full model. \\
        \textbf{Bold} and \underline{underline} indicates the top and second highest success rates in each column.
    }
\end{table}

\subsubsection{Hyperparameter Sensitivity} 
\par
The default configuration uses a handover point of $r=0.4$, a two-layer Mamba execution memory module with width 1024, and a four-layer PAE. Each configuration in Table~\ref{tab:parameter_ablation} varies one hyperparameter while keeping the others fixed.

\par
\noindent\textbf{Expert handover.} The handover point $r$ determines the fraction of the ten denoising steps assigned to the memory-conditioned PAE before the frozen AE takes over. Ratios of $0.3$, $0.4$, $0.5$, and $0.6$ yield average success rates of 48.3\%, 55.0\%, 45.0\%, and 47.2\%, respectively. The default $r=0.4$ performs best on all three tasks among the tested ratios. A smaller ratio allocates fewer steps to memory-conditioned denoising, while a larger ratio leaves fewer refinement steps to the frozen AE. These results favor an intermediate allocation of four PAE steps followed by six AE denoising steps.

\par
\noindent\textbf{Execution memory capacity.} Increasing Mamba depth from two to four reduces average success from 55.0\% to 43.3\%. Increasing its width from 1024 to 1536 raises the average to 55.6\%, although success on \texttt{T6} decreases from 80.0\% to 76.7\%. Given this small numerical gain of 0.6 percentage points, we retain the narrower, two-layer configuration, which achieves a similar average success rate with less computation burden. 

\par
\noindent\textbf{PAE capacity.} Reducing PAE depth from four to two lowers average success from 55.0\% to 43.9\%. Increasing the depth to six improves \texttt{T8} from 45.0\% to 55.0\%, but reduces success on \texttt{T6} and \texttt{T7}, yielding an average of 53.9\%. We therefore retain the four-layer PAE, which achieves the highest average success rate among the tested depths. The capacity experiments indicate that increasing model size does not consistently improve performance in these settings.

\begin{table}[h]
    \centering
    \caption{Hyperparameter sensitivity on \texttt{T6}--\texttt{T8}
    (Success Rate, \%).}
    \label{tab:parameter_ablation}
    \setlength{\tabcolsep}{2pt}
    \begin{tabular*}{\columnwidth}{
        @{\extracolsep{\fill}}lccccc@{}
    }
        \toprule
        \textbf{Parameter} & \textbf{Default} $\rightarrow$ \textbf{Tested} & \texttt{T6} & \texttt{T7} & \texttt{T8} & \textbf{Avg.} \\
        \midrule
        Handover point  & $0.4 \rightarrow 0.3$  & 65.0 & \underline{36.7} & 43.3 & 48.3 \\
         & $0.4 \rightarrow 0.5$  & 66.7 & 25.0 & 43.3 & 45.0 \\
        & $0.4 \rightarrow 0.6$  & 71.7 & 28.3 & 41.7 & 47.2 \\
        \midrule
        Mamba depth    & $2 \rightarrow 4$ & 58.3 & 25.0 & 46.7 & 43.3 \\
        Mamba width    & $1024 \rightarrow 1536$  & \underline{76.7} & \textbf{40.0} & \underline{50.0} & \textbf{55.6} \\
        \midrule
        PAE depth      & $4 \rightarrow 2$ & 70.0 & 21.7 & 40.0 & 43.9 \\
        & $4 \rightarrow 6$ & 71.7 & 35.0 & \textbf{55.0} & 53.9 \\
        \midrule
        \textbf{Default cfg.}  & -- & \textbf{80.0} & \textbf{40.0} & 45.0 & \underline{55.0} \\
        \bottomrule
    \end{tabular*}

    \vspace{3pt}
    \parbox{\columnwidth}{
        \scriptsize
        \textit{Default configuration:} $r=0.4$, Mamba width/depth $=1024/2$, and PAE depth $=4$.
        Each ablation changes only the indicated parameter from the default configuration.
        \textbf{Bold} and \underline{underline} indicates the top and second highest success rates in each column.
    }
\end{table}

\subsection{Real-World Results}
\label{sec:exp-real}
\par
The real-world tasks extend the simulation evaluation by requiring explicit transitions after weighing or a prescribed number of repetitions. In \texttt{R1}, \texttt{R3}, and \texttt{R4}, successful completion requires both performing the specified repetitions and switching to a distinct terminal action at the appropriate stage. Indefinite repetition alone therefore cannot satisfy the task requirements.

\par
\begin{figure}[bt!]
  \centering
  \includegraphics[width=0.85\columnwidth]{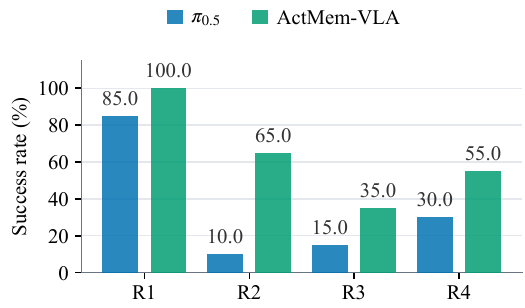}
  \caption{Success rates on four real-world manipulation tasks.
   \texttt{R1:} Two-Cycle Bowl Return.
  \texttt{R2:} Mango Weighing.
  \texttt{R3:} Triple Button Press.
  \texttt{R4:} Three-Cycle Bowl Return.
  }
  \label{fig:real_world_success}
\end{figure}

As shown in Fig.~\ref{fig:real_world_success}, \method{} outperforms the memoryless $\pi_{0.5}$ baseline on all four tasks. Success increases from 85\% to 100\% on \texttt{R1}, from 10\% to 65\% on \texttt{R2}, from 15\% to 35\% on \texttt{R3}, and from 30\% to 55\% on \texttt{R4}. With 20 trials per task, \method{} succeeds in 51 of 80 trials, compared with 28 of 80 for $\pi_{0.5}$. The average success rate thus rises from 35.0\% to 63.75\%, an improvement of 28.8 percentage points.

\par
The largest gain occurs on \texttt{R2}, where the policy must distinguish between similar robot-object configurations before and after weighing. This improvement is consistent with better discrimination between task stages that cannot be reliably identified from the current observation alone. Gains on \texttt{R1}, \texttt{R3}, and \texttt{R4} further support the usefulness of execution history for tasks requiring repetition tracking followed by a change in action. Reliable execution of the three-repetition tasks remains challenging, with success rates of 35\% on \texttt{R3} and 55\% on \texttt{R4}, compared with 100\% on the two-cycle \texttt{R1} task (Fig.~\ref{fig:real_world_success}). The improvements across all four tasks support the effectiveness of \method{} for real-world manipulation requiring both repetition tracking and appropriate stage transitions.

\section{Conclusion}
\par
In this work, we presented \method{}, a memory-augmented VLA framework designed to address task-progress ambiguity in long-horizon manipulation. \method{} leverages executed-action history as compact progress-aware memory, encoded by Mamba to condition a lightweight PreAction Expert (PAE). The PAE steers the early, high-noise denoising process toward history-consistent actions before handing the intermediate action state to the frozen Action Expert (AE) for fine-grained refinement. By training only the memory module and PAE while keeping the fine-tuned base VLA entirely frozen, \method{} enables parameter-efficient temporal adaptation without compromising the learned action-generation capabilities of the base policy. This design provides a practical pathway for equipping existing VLAs with temporal memory to support more coherent and reliable long-horizon manipulation. Experiments on ten simulated tasks and four real-world tasks demonstrate improved overall task completion. Under demonstration-based task-specific step limits, \method{} achieves 80.8\% average success in simulation, exceeding $\pi_{0.5}$ and MemoryVLA by 15.6 and 31.3 percentage points, respectively. In real-world experiments, it improves success on all four tasks and raises the average from 35.0\% to 63.75\% relative to $\pi_{0.5}$. 

These results support its effectiveness on tasks requiring stage discrimination, repetition tracking, and transitions to subsequent actions. Component ablations support the combined use of execution memory and a trainable early-denoising expert, while revealing task-dependent benefits. The proposed separation of trainable history-conditioned action generation from frozen policy refinement provides a practical approach to incorporating execution memory into existing VLAs.



\bibliographystyle{IEEEtran}
\bibliography{IEEEabrv, references}

\end{document}